\documentclass{article}

\usepackage[round]{natbib}

\usepackage[preprint]{neurips_2024}

\usepackage[utf8]{inputenc} % allow utf-8 input
\usepackage[T1]{fontenc}    % use 8-bit T1 fonts
\usepackage{hyperref}       % hyperlinks
\usepackage{url}            % simple URL typesetting
\usepackage{booktabs}       % professional-quality tables
\usepackage{amsfonts}       % blackboard math symbols
\usepackage{nicefrac}       % compact symbols for 1/2, etc.
\usepackage{microtype}      % microtypography
\usepackage{xcolor}         % colors

\usepackage{graphicx}
\usepackage{amsmath}

\title{Zero-shot Shark Tracking and Biometrics from Aerial Imagery}

\author{%
  Chinmay K. Lalgudi \\
  Stanford University \\
  Stanford, CA, USA \\
  \texttt{clalgudi@stanford.edu} \\
  \And 
  Mark E. Leone \\
  Stanford University \\
  Stanford, CA, USA \\
  \And 
  Jaden V. Clark \\
  Stanford University \\
  Stanford, CA, USA \\
  \And 
  Sergio Madrigal-Mora \\
  Flinders University \\
  Adelaide, SA, Australia \\
  \And 
  Mario Espinoza \\
  Centro de Investigación en Ciencias del Mar y Limnología, \\ Universidad de Costa Rica, San José, Costa Rica \\
}

\begin{document}

\maketitle

\begin{abstract}
1. The recent widespread adoption of drones for studying marine animals provides opportunities for deriving biological information from aerial imagery. The large scale of imagery data acquired from drones is well suited for machine learning (ML) analysis. Development of ML models for analyzing marine animal aerial imagery has followed the classical paradigm of training, testing, and deploying a new model for each dataset, requiring significant time, human effort, and ML expertise.

2. We introduce Frame Level ALIgment and tRacking (FLAIR), which leverages the video understanding of Segment Anything Model 2 (SAM2) and the vision-language capabilities of Contrastive Language-Image Pre-training (CLIP). FLAIR takes a drone video as input and outputs segmentation masks of the species of interest across the video. Notably, FLAIR leverages a \textit{zero-shot} approach, eliminating the need for labeled data, training a new model, or fine-tuning an existing model to generalize to other species.

3. With a dataset of 18,000 drone images of Pacific nurse sharks, we trained state-of-the-art object detection models to compare against FLAIR. We show that FLAIR massively outperforms these object detectors and performs competitively against two human-in-the-loop methods for prompting SAM2, achieving a Dice score of 0.81. FLAIR readily generalizes to other shark species without additional human effort and can be combined with novel heuristics to automatically extract relevant information including length and tailbeat frequency. 

4. FLAIR has significant potential to accelerate aerial imagery analysis workflows, requiring markedly less human effort and expertise than traditional machine learning workflows, while achieving superior accuracy. By reducing the effort required for aerial imagery analysis, FLAIR allows scientists to spend more time interpreting results and deriving insights about marine ecosystems.
\end{abstract}

\section{Introduction}
\label{sec:intro}

Large marine animals, such as sharks, influence ecosystems through a variety of mechanisms as predators, nutrient transporters, and even prey, thus maintaining the balance of marine food webs, regulating species populations, and promoting biodiversity \citep{heupel2014sizing, heithaus2022advances, dedman2024ecological}. Unfortunately, overfishing and other anthropogenic threats have greatly reduced shark populations, altering their ecological roles and effects on ecosystems \citep{stevens2000effects, ferretti2010patterns, myers2007cascading}. As a result, long-term monitoring of sharks and other large marine animals is key to understanding how animal populations are responding to human impacts and potential environmental shifts caused by climate change \citep{Brock2013, Torres2022}. However, studying sharks is challenging due to their elusive nature, vast ranges, and the inherent difficulties in observing their interactions in marine environments \citep{jorgensen2022emergent}. Thus, new technology must be developed to understand both their large-scale and precise movement ecology.

Researchers often use satellite or acoustic telemetry to study shark behavior over multiple spatial and temporal scales, but tagging requires significant human effort and may be detrimental to the well-being of tagged animals \citep{kohler2001shark, matley2024long}. More recently, baited remote underwater video (BRUV) systems have been used to capture visual information and collect imagery \citep{white2013application}. However, BRUVs are limited to capturing localized behavior and are dependent on sharks maintaining proximity to the deployment site.

The use of unmanned aerial vehicle (UAV) systems is emerging as a promising approach for the non-invasive study of volitional marine animal behavior and biometrics \citep{gray2019drones, hodgson2013unmanned, ramos2022drone, torres2020morphometrix, Torres2022}. Aerial imagery can be used to compute biometrics such as length, body condition, tailbeat frequency, and relative velocities, which can provide key information about animal health \citep{bierlich2024automated}, swimming kinematics \citep{porter2020volitional}, and predator-prey interactions \citep{hansen2022mechanisms}.  There has been significant effort towards developing UAV systems for studying wildlife \citep{butcher2021drone, shah2020multidrone, hodgson2018drones, jadhav2024reinforcement}, including path-planning across varying environmental conditions \citep{shah2020multidrone, clark2024online} and exploring the efficacy of sensor modalities for wildlife detection \citep{saunders2022radio, beaver2020evaluating}. In this work, we focus on developing automated systems for analyzing existing drone imagery. 

\subsection{Deep Learning for Marine Ecology}

Deep learning, the process of training large artificial neural networks to learn complex functions from data, has important applications to the study of aerial wildlife imagery \citep{lecun2015deep}. Previous works have used deep learning for analysis of aerial imagery; however, they often rely on specialized object detection models \citep{carrio2017review, eikelboom2019improving}. 

Traditionally, object detection networks have relied on
Convolutional Neural Networks (CNN) architectures and their variants \citep{li2021survey,cnnreview, girshick2015fast, ren2015faster}. 
Perhaps the most popular object detection model with a CNN backbone, You Only Look Once (YOLO) \citep{Jocher_Ultralytics_YOLO_2023}, is specialized for inference speed, treating object detection as a regression problem. By predicting classes and bounding boxes in a single pass, YOLO models are more suitable for real-time applications. More recently, the Detection Transformer (DETR) \citep{carion2020end} has achieved state of the art results by integrating transformers into the encoder and decoder of the model. This eliminates the need for many hand-engineered components by doing direct set prediction of object classes and bounding boxes. DETR leverages this global context to achieve impressive results thar rival  non-transformer architectures. This is particularly beneficial in scenarios discussed in this work, where multiple objects are in close proximity, such as large groups of sharks, or partially occluded by factors like camera glare or high turbidity.

Some studies have proposed using object detection models to track sharks in aerial imagery \citep{pavithraetal, zhaoetal, butcher2021drone}. One of the earliest efforts to apply neural networks to aerial imagery of marine life consisted of training a vanilla CNN for sea turtle detection \citep{gray2019convolutional}, and subsequent works have used transfer learning (fine-tuning a pre-trained CNN), improving model performance \citep{gray2019convolutional, desgarnier2022putting, sharma2018shark}. Alternatively, some approaches train models by directly segmenting marine animals and objects. \citet{pavithraetal} propose a novel hybrid architecture called SwinConvMixerUNet for underwater image segmentation, leveraging the Swin Transformer's ability to capture spatial information and the ConvMixer's channel-mixing capabilities to enhance feature extraction and segmentation accuracy.

Unfortunately, standard object detection and segmentation models require large datasets of high-quality human-annotated data to train models. Furthermore, they often do not generalize well, performing poorly when the inference data distribution differs from the model's training data \citep{koh2021wilds}. In contrast to conventional approaches, we leverage foundation models (i.e. large deep learning models trained on internet-scale datasets) for marine animal tracking and biometric analysis from aerial imagery \citep{bommasani2021opportunities}. The key advantage of using pre-trained foundation models is that they can be deployed \textit{zero-shot}. Thus, they do not require dataset curation or training to adapt to new data, and require significantly less human effort and expertise to use. There is a notable lack of adoption of foundation models for the study of marine animals from UAV imagery, with the singular exception being an automated pipeline using Segment Anything Model for surveying whale length and body condition \citep{bierlich2024automated}.  

In this work, we explore methods for automatically computing segmentation masks and downstream biometrics for sharks using Segment Anything Model 2 (SAM 2), a pre-trained foundation model for promptable image and video segmentation \citep{ravi2024sam}, and Contrastive Language-Image-Pretraining (CLIP), an approach for learning shared representations between natural language and pixels \citep{radford2021learning}.  We present a new method, Frame Level AlIgnment and tRacking (FLAIR), that uses CLIP and SAM 2 to generate accurate segmentations for several shark species across different environments, leveraging a zero-shot approach that eliminates the need for annotating data, training, or fine-tuning. These segmentation masks can be used to compute tailbeat frequency, length, mass, velocities, and other downstream biometrics for arbitrary shark species. 

We test our approach predominantly on a dataset of the Pacific nurse shark (\textit{Ginglymostoma unami}) from Santa Elena Bay (North Pacific coast of Costa Rica), demonstrating how our method can be used to help better understand the movement ecology of an endangered, data-poor species \citep{madrigal2024long}. We compare segmentation accuracy of FLAIR with multiple approaches, including prompting SAM 2 with a human in the loop, as well as prompting SAM 2 with state-of-the-art object detection models. 

Our study suggests that FLAIR is capable of generalizing to other species and we show that FLAIR segmentations can be used to measure biometrics including length and tailbeat frequency. Notably, FLAIR does not require any training or fine-tuning to generalize to other species, highlighting its potential applicability across diverse ecosystems. 

% Faster RCNN architectures are composed of a feature extractor backbone (typically a pre-trained CNN) followed by a region proposal network to predict object boundaries and their "objectness" scores, and finally region-of-interest (ROI) pooling to extract uniform-sized features.
\begin{figure*}[h!]
  \centering
  \includegraphics[width=0.75\linewidth]{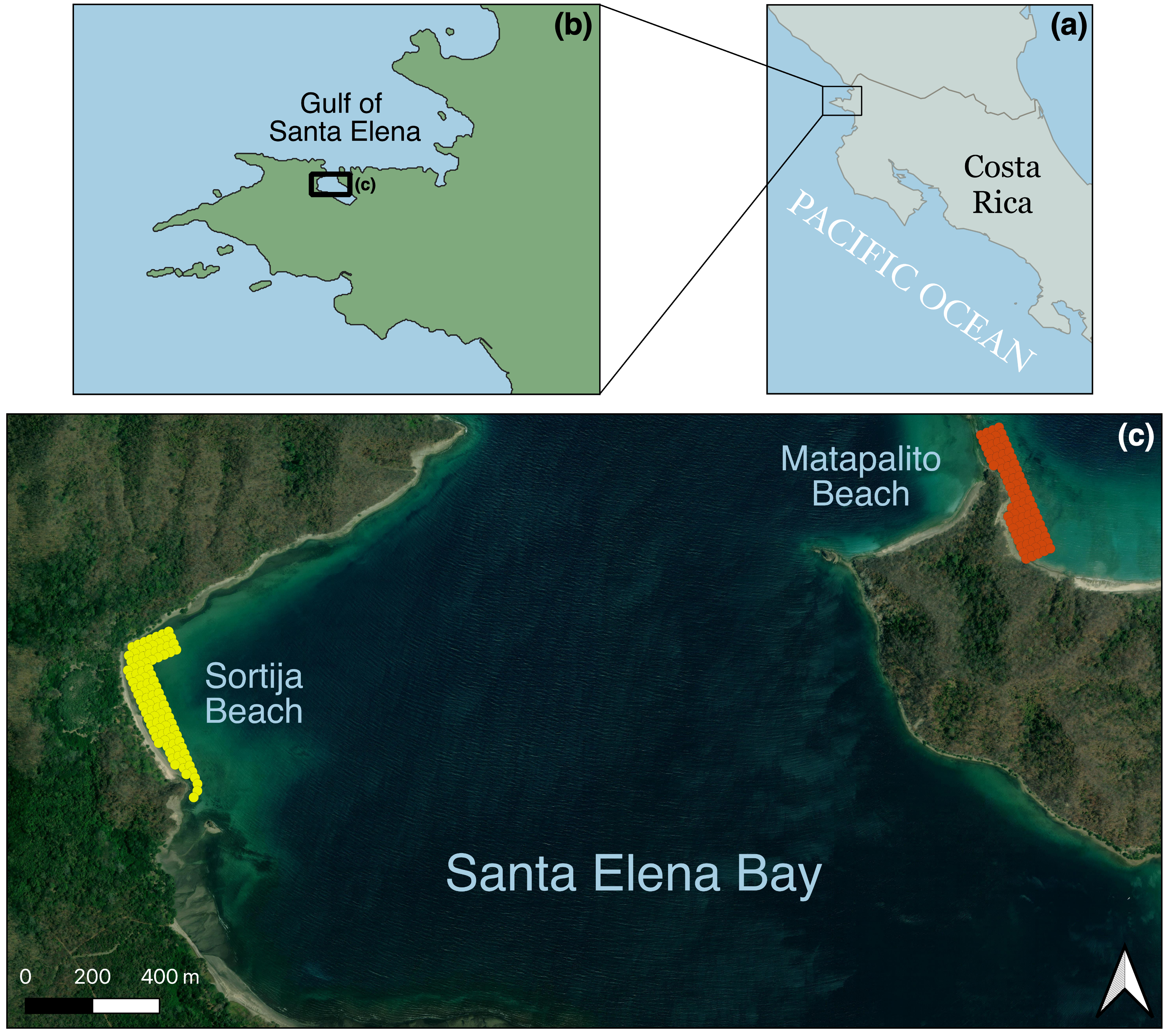}
  \caption{A map of the sites where drone video data for the Pacific nurse shark dataset was collected in Santa Elena Bay, Guanacaste, Costa Rica (a). Inset map in (b) shows the Santa Elena Bay study area (c), in which the yellow and orange dots represent the pre-planned flight path locations at Sortija Beach and Matapalito Beach, respectively.}
  \label{fig:map}
\end{figure*}

\section{Materials and Methods}
\subsection{Dataset}

The Pacific nurse shark imagery was collected from two field sites (Matapalito Beach and Sortija Beach) in the coastal waters of the Eastern Tropical Pacific Ocean, in Santa Elena Bay, Costa Rica  (Fig. \ref{fig:map}a,  b). The dataset was collected from 2022-2024, over a period of 23 months, with varying water visibility (turbidity), illumination, and wind/wave conditions. Images were collected at each site by flying a DJI Mavic 2 drone on a pre-programmed path, recording a continuous video at 30FPS and 3840×2160 resolution. The drone stopped at predetermined waypoints for 3 seconds each (Fig. \ref{fig:map}c, yellow and orange dots).

\begin{figure*}[h!]
  \centering
  \includegraphics[width=\linewidth]{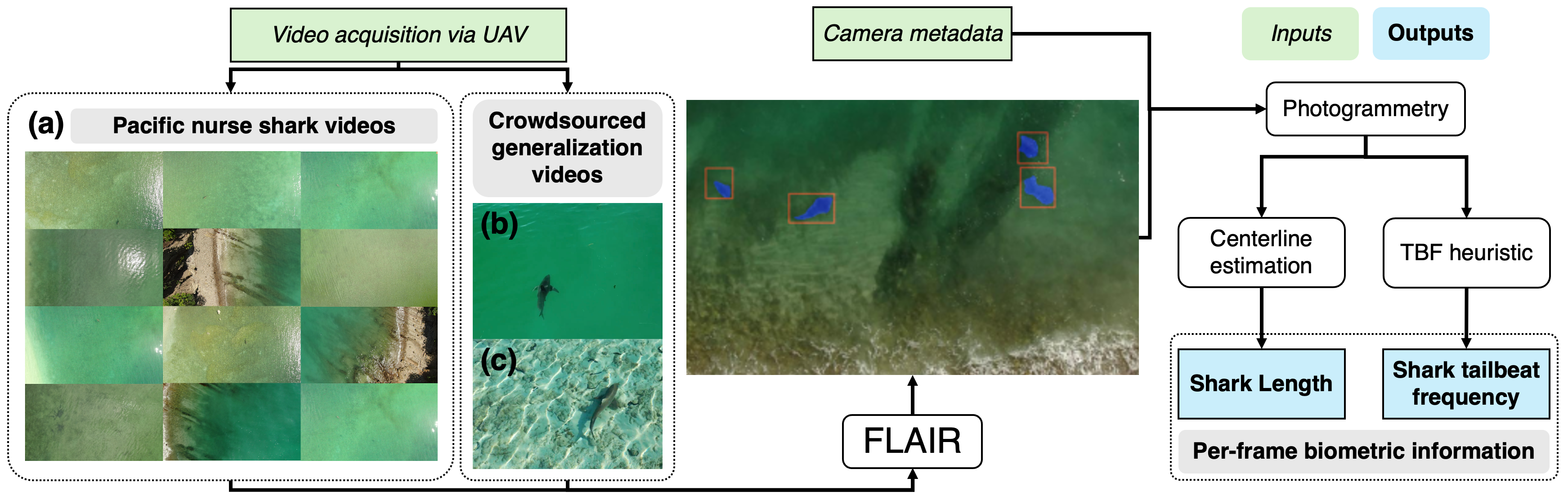}
  \caption{Automated biometrics workflow with FLAIR. Inputs are italicized and in green, outputs are bolded and in blue. (a) Representative frames from the Pacific nurse shark UAV video dataset. Tests of generalization of FLAIR beyond our study site and selected species was conducted with crowdsourced videos of a white shark (b) and a blacktip reef shark (c).}
  \label{fig:datast}
\end{figure*}

More than 6 hours of video was recorded in total, during 60 drone surveys, resulting in 648,000 total frames captured. To our knowledge, this is one of the largest open-access datasets of nearshore shark aerial drone imagery.. For object detection, the dataset was pruned to include a diverse set of 7 videos from the two sites, across varying tidal, turbidity, water surface glare, and wave conditions. Ground truth bounding boxes were added for each Pacific nurse shark with the Computer Vision Annotation Tool (CVAT) \citep{cvat}. The images from 2 videos were completely separated from the rest of the dataset, as a test for generalization. The rest of the dataset was time-blocked such that each 45 adjacent frames of a video were held together when the data was split into training, validation, and test sets. This time-blocking was done to minimize occurrences of consecutive frames being present in the training and test set, which would artificially inflate models' performance metrics. Sharks were present in diverse conditions across videos and time segments, including variable turbidity and substrate type  (Fig. \ref{fig:datast}a). Prior to object detection model training, images were rescaled from \(3840 \times 2160\) pixels to \(1080 \times 1080\) pixels. Standard data augmentation techniques, including random rotation, brightness, and hue adjustments, were also applied.

Excluding the images from the two videos held for generalization testing, our object detection dataset contained 9,200 unique positive examples (images containing sharks) and a corresponding 27,000 unique negative examples (images containing no sharks). We selected 9,200 negative examples to remain in the dataset, so that our dataset contained an equal ratio of images with and without sharks. The dataset was then randomly split into training, validation, and test sets, in a [80-10-10] ratio, maintaining the images in these 45-frame time blocks. This resulting dataset used for training and testing the object detectors contains 18,400 total images.

The object detectors were tested on an internal test set from the 5 training videos, as well as the 2 holdout videos. The other approaches—Per-frame Prompting, HiL-Tracking, and FLAIR—did not require training, thus they were tested on all 7 videos. For computational efficiency, smaller sub-videos containing sharks were used as input into these pipelines, each ranging from 20 to 80 seconds in length. This dataset was used for precision and recall metric comparisons across all methods. A random sampling of 100 frames from the 2 holdout videos and 200 frames from the other 5 videos were annotated for segmentation ground truth masks and biometrics using CVAT, as described in Section \ref{sec:biomet_methods}. These ground truth masks were used to compute the accuracy of segmentation across methods.

To test the generalization abilities of FLAIR, two UAV videos licensed under Creative Commons Attribution were acquired from YouTube—one of a white shark (\textit{Carcharodon carcharias}) in Southern California, USA, and one of a blacktip reef shark (\textit{Carcharhinus melanopterus}) in the Great Barrier Reef, Australia (Fig. \ref{fig:datast}b,c). Full attribution to the original creators of these videos is provided in the Supporting Information section. A smaller dataset was built from these blacktip and white shark videos, sampling 25 random frames from each of the two videos. Similarly, segmentation mask ground truths and biometrics were manually annotated to test the generalizability of FLAIR across diverse aerial footage.

\subsection{Baselines}
\subsubsection{Per-frame Prompting}

We evaluated baseline methods for object segmentation traditionally used in the field, with one of the most popular approaches being per-frame prompting. This method requires a human annotator to label bounding boxes for sharks in every frame of a video, which is effective for accurate object detection but extremely laborious (Fig. \ref{fig:methods_summary}a). We hand-labeled bounding boxes in 18,400 frames across 5 videos, prompting each frame and corresponding bounding boxes into SAM 2 Image Prediction to get masks for sharks in every image. Notably, the segmentation aspect of this method does not utilize video understanding, as each frame was segmented as an individual image.

\subsubsection{Object Detection Models}

We also focused on evaluating a series of object detection models, predominantly leveraging the Detectron2 model library \citep{wu2019detectron2}. First, we employed several RCNN model architectures, two-stage detectors optimized for accuracy (as opposed to inference speed like many one-stage detectors) \citep{girshick2015fast}. We trained several RCNNs with various backbones (ResNet and ResNeXt architectures) and feature pyramid networks (FPN) to enhance the effect of objects of various scales in images. All of these models were pre-trained on the Imagenet dataset \citep{deng2009imagenet}.

\begin{figure*}[h!]
  \centering
  \includegraphics[width=\linewidth]{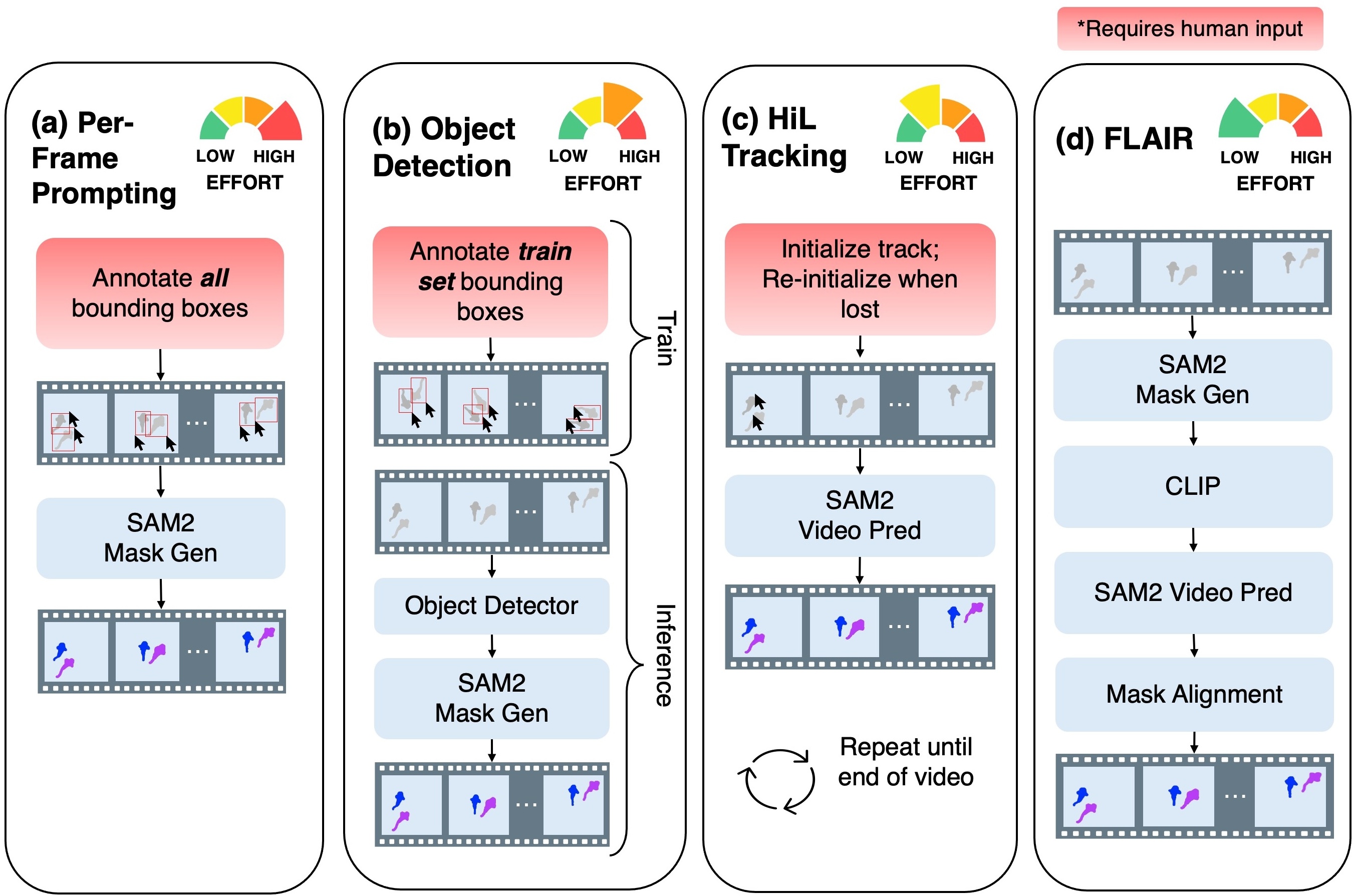}
  \caption{Summary of segmentation methods compared in our experiments for Per-frame Prompting with SAM 2 Mask Generation (a), Object Detection methods paired with SAM 2 Mask Generation (b), Human-in-the-Loop Tracking with SAM 2 Video Prediction (c), and FLAIR (d). Steps requiring human input/effort are shown in red. Effort is shown in the upper right of each figure. }
  \label{fig:methods_summary}
\end{figure*}

We also trained the DETR architecture for shark detection \citep{carion2020end}. Finally, we trained a YOLOv8 model pre-trained on Imagenet \citep{Jocher_Ultralytics_YOLO_2023}. The pre-trained medium YOLOv8 model was trained for 40 epochs using default hyperparameters (until loss converged). DETR was trained for 120 epochs, and the rest of Detectron models were trained for 40 epochs each - all with a maximum of 100 objects detected per frame. Each frame, SAM 2 was prompted with bounding boxes output from the object detector to generate segmentation masks (Fig. \ref{fig:methods_summary}b). 

\subsection{Human-in-the-Loop (HiL) Tracking}
In human-in-the-loop tracking, a human manually annotates a bounding box in the first frame a shark is identified, and then  SAM 2 is used to track the segmentation through the remainder of the video (until it was lost) (Fig. \ref{fig:methods_summary}c). When the track is lost, the annotator re-initializes the segmentation track with a bounding box. CVAT was used for bounding box initialization.

\subsection{FLAIR}
\begin{figure*}[h!]
  \centering
  \includegraphics[width=\linewidth]{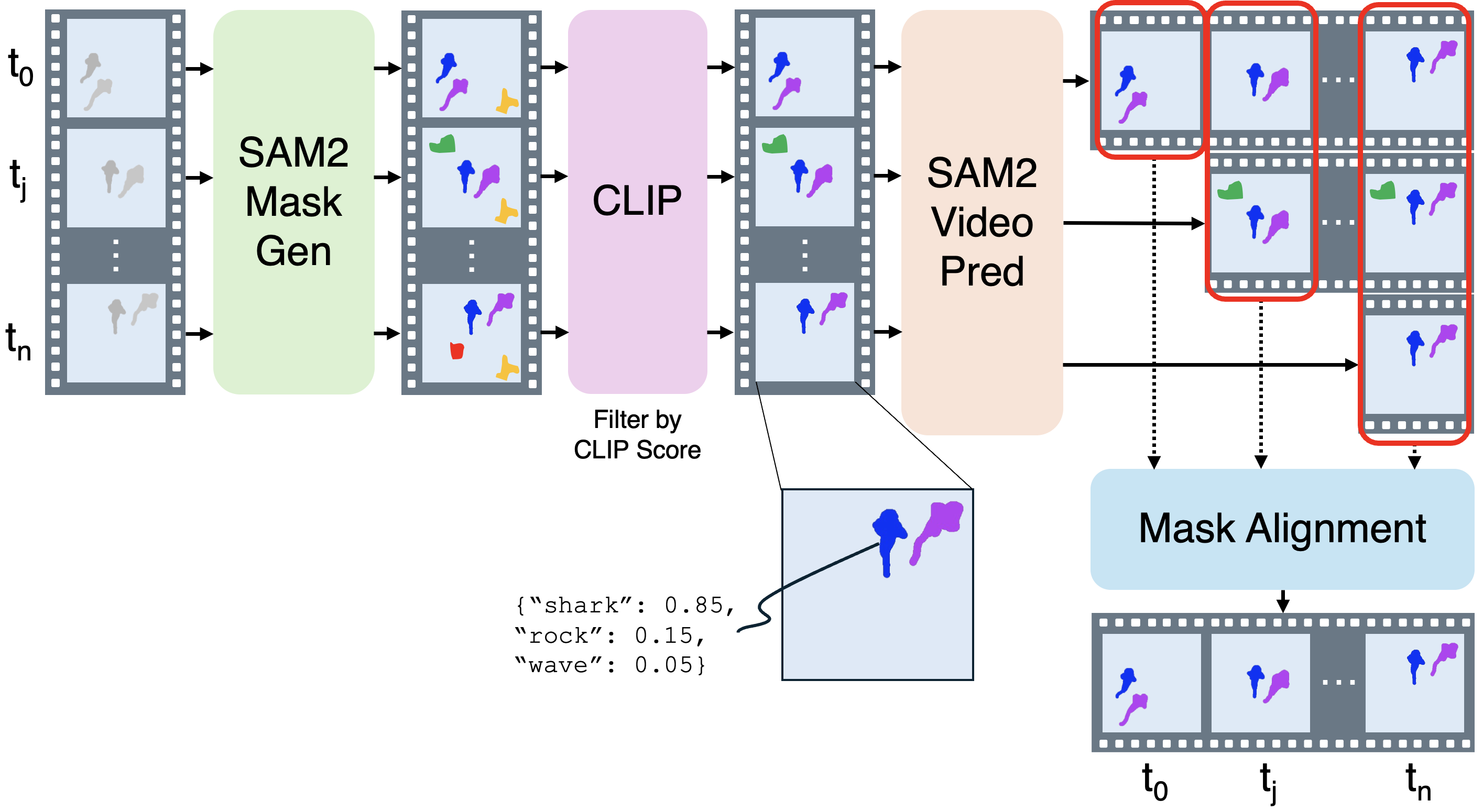}
  \caption{Detailed overview of FLAIR architecture. Input video is passed into SAM 2 Mask Generation to segment all objects, which are then filtered by CLIP score given language prompts. Candidate masks are propagated through the video and aligned to eliminate false positives, resulting in accurate tracking of objects of interest. }
  \label{fig:flair}
\end{figure*}

Our proposed method, FLAIR, is a fully autonomous framework for object tracking—integrating frame-level alignment and video understanding with language prompts, as illustrated in Fig. \ref{fig:flair}. First, individual frames of the video are sampled at a uniform time interval and are passed into the SAM 2 Automatic Mask Generator. In this work, time intervals of 30 frames (1 sec) were used. SAM 2 Automatic Mask Generator generates masks for all possible objects in the image by sampling single-point inputs in a grid, filtering and de-replicating candidate masks, and performing further processing for improved quality. Bounding boxes are generated for each mask in the frame and passed into CLIP, along with prompts that were fine-tuned for this task. The specific prompts used are included in Supporting Information. The prompts were held constant across all of the nurse shark videos as well as the white and blacktip reef shark videos, highlighting the generalizability of this method. Each bounding box is assigned a probability associated with each prompt, and all bounding boxes with a probability assigned to the shark prompt greater than 0.95 are kept as candidate sharks. Frames with their corresponding candidate bounding boxes are then initialized in SAM 2 Video Prediction to track the candidate sharks through the remainder of the video. SAM 2 Video Prediction is provided with the candidate bounding box from its respective frame but is prompted to track this object from the beginning of the video, allowing tracks to propagate to previous frames. This video tracking is performed at every time interval, initialized with the candidate shark bounding boxes in that frame. As the mask is tracked through the video, if the mask overlaps with another candidate mask from another time interval track with an IOU greater than 0.7, the masks are determined to be aligned and thus declared a true positive mask for a shark. For example, we see in Fig. \ref{fig:flair} that the false positive mask (in green) was propagated through the second time interval track, but was not present in any other tracks. Thus, it was not classified as a true shark mask during alignment. FLAIR was applied to analyze the dataset using a single NVIDIA A100 GPU. 

The advantage of FLAIR comes from the fact that presumably, there are a small number of false positives that are still recognized as candidate sharks by CLIP. Although these false positives can be propagated through the video from a single frame, the same candidate shark mask will likely be absent at future time intervals. Thus, the false positives identified in individual frames will not be aligned across tracks initialized at other frames—increasing robustness and generalizability across diverse sets of videos.

\subsection{Biometrics Measurements}
    \label{sec:biomet_methods}
Segmentations of marine animals can be used for further downstream tasks including the calculation of biometrics \citep{bierlich2024automated, gray2019drones}. Tailbeat frequency (TBF) and length were computed from FLAIR-predicted masks and compared against manual calculations for a video of a Pacific nurse shark from our dataset, and open-access videos of a blacktip reef shark and a white shark. Manual measurements for length were computed in pixels using CVAT line tool for a random sample of frames for each of the videos \cite{cvat}. TBF was manually measured by observing a complete tail beat starting from the equilibrium position aligned with the center line of the shark and returning to this equilibrium position, recording the frame numbers for each full cycle.

To estimate the length of shark along its centerline, we first obtained the segmentation mask predicted by FLAIR. The mask is skeletonized using Zhang's method \citep{zhang1984fast} by thinning the image around the boundaries over successive passes, eventually obtaining the skeleton of the mask. The total length of the skeleton was calculated by traversing each point on the backbone, extending the skeleton to include the distal ends of the mask.

The length \( L_{\text{pixels}} \) obtained from the mask, measured in pixels, was converted to length in meters \( L_{\text{m}} \) using the following function, modified from \citep{torres2020morphometrix}:
\begin{equation}
L_{\text{m}} = \left( \frac{S_w}{I_w} \times \frac{A + D}{F} \right) \times L_{\text{pixels}}
\label{eq:lengthconv}
\end{equation}
where \( S_w \) is the sensor width in mm (13.2 mm for a 1" CMOS sensor), \( I_w \) is the image width in pixels (1920 pixels for 1920x1080p resolution), \( A \) is the altitude of the drone in meters (37 meters), \( D \) is the depth of the shark in meters, and \( F \) is the focal length of the camera in mm (28 mm). We assumed a constant depth \( D \) of 1.5 m for this analysis, since Pacific nurse sharks were typically observed at approximately 1.5 m depth in snorkeling surveys in Santa Elena Bay. For the two open-access videos of the blacktip reef shark and white shark, the exact camera, sensor, and drone altitude metadata were not available. Thus, we held these values constant across all 3 videos and only varied the drone altitude, estimating this quantity from the videos such that the resulting shark length was a typical value for that species. Therefore, the calculations for the blacktip reef shark and white shark should only be used as an example of the capabilities of this workflow, and not to draw scientific conclusions.

To calculate TBF, the two furthest points in the mask were calculated, with the closer point to the center of mass (COM) of the mask labeled as the head and the farther point deemed the tail. The center line of the shark was calculated as the vector from the head to the COM, and the vector from the COM to the tail was projected onto this center line, giving the orthogonal distance from the tail to the center line. This distance is calculated for every mask at each frame in the video and plotted across time to give a sinusoidal curve. Smoothing is performed by applying a Savitzky-Golay filter across the distances. Then, the points where the smoothed curve crossed the center line (y = 0) were calculated, with local minima and maxima also identified. To eliminate small deviations in the distance that persisted after smoothing, we assumed that meaningful crossings across the center line occurred between local extrema. Only keeping these crossings, we then calculated the intervals for tail beats as every other crossing that was aligned with the equilibrium center line, proving both conceptually and empirically to be a strong detector of TBF. To compare predicted tail beat intervals with manually-measured tail beat intervals, we moved a sliding window of 5 seconds across the video in steps of 0.5 seconds, calculating the fractional number of tail beats for each window. This  allows for both comparison between manual and predicted TBF and observing TBF across time, providing potential insight into shark kinematics and energetics.

\section{Results}
\subsection{Object Detection}

 We tested our suite of models on an internal test dataset to assess the models' ability to learn representations, as well as two previously-unseen external holdout videos to assess the generalizability of the models. Testing the models on both datasets gauges potential overfitting on previously-seen data, while also evaluating for practical usage on drone imagery from new distributions. Among the models, YOLOv8 and DETR performed the best on the internal holdout test set across all metrics, including mean Average Precision (mAP) and mean Average Recall (mAR) from IOU thresholds between 0.5 to 0.95 and Average Precision and Average Recall at IOU thresholds of 0.35 and 0.5. The exact precision and recall values are shown in Table S1. YOLO had the highest mAP as well as the highest mAR, with DETR performing slightly better at lower IOU thresholds. Both of these performed better than the trained Faster R-CNN models with Feature Pyramid Network and Dilated-C5 backbones. It is important to note that although mAP averaged between 0.5 and 0.95 are traditionally used as performance metrics for object detection models, we found that it is not necessary for predicted bounding boxes to have a high IOU with ground truth boxes in order to obtain accurate results on downstream biometrics methods. Thus, although both YOLO and DETR had a relatively low mAP and mAR of 0.63 and 0.70, performance at a lower IOU threshold of 0.35 was significantly higher—with DETR achieving a near-perfect AP and AR of 0.94 and 0.99, respectively.
 
 As seen in Table S2, YOLOv8 and DETR had high accuracy and recall at lower IOU thresholds on holdout Video 1, but both performed extremely poorly on holdout Video 2, with mean AR at IOU threshold of 0.1 being 0.03 and 0 respectively. FLAIR had similar AP and AR for the first holdout video, but had significantly better performance in Video 2, with an AP and AR at IOU of 0.1 equaling 0.49 and 0.95 respectively.

\begin{table*}

\label{tab:all_methods_vids} % Label added here
\resizebox{\textwidth}{!}{%
\begin{tabular}{|c|c|c|c|c|c|c|c|c|c|c|c|c|}
\hline
\multicolumn{1}{|c|}{} & \multicolumn{2}{c|}{\textbf{Holdout Videos}} & \multicolumn{5}{c|}{\textbf{Training Videos}} \\ \hline
\textbf{Model} & \textbf{Video 1} & \textbf{Video 2} & \textbf{Video 3} & \textbf{Video 4} & \textbf{Video 5} & \textbf{Video 6} & \textbf{Video 7} \\ \hline
Per-frame Prompting + SAM 2 & 0.756 & 0.780  & 0.852 & 0.879 & 0.669 & 0.845 & 0.737 \\ \hline
YOLOv8 + SAM 2 & 0.595 & 0.014 & --- & --- & --- & --- & ---  \\ \hline
DETR + SAM 2 & 0.700 & 0 & --- & --- & --- & --- & ---  \\ \hline
HiL-Tracking + SAM 2 Video & 0.734 & 0.839 & 0.847 & 0.850 & 0.838 & 0.850 & 0.852 \\ \hline
FLAIR & 0.740 & 0.837 & 0.849 & 0.860 & 0.839 & 0.722 & 0.853   \\ \hline
\end{tabular}
}
\caption{Performance metrics (Dice Score) of segmentation masks predicted by GT + SAM 2, YOLOv8 + SAM 2, DETR + SAM 2, and FLAIR on holdout videos.}
\end{table*}

\begin{figure*}[h!]
  \centering
  \includegraphics[width=\linewidth]{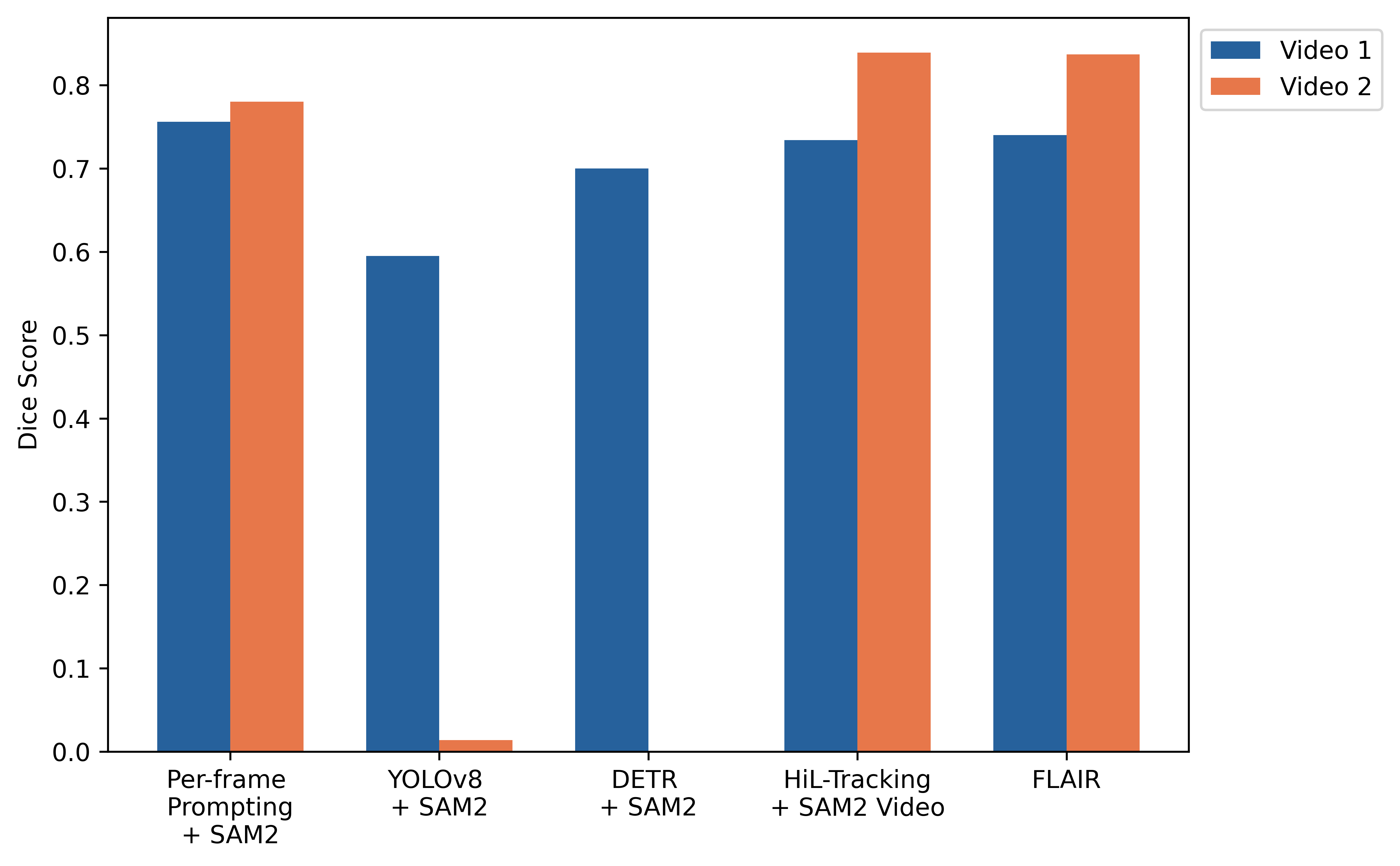}
  \caption{Dice score comparison of Per-frame Prompting + SAM 2, YOLOv8 + SAM 2, DETR + SAM 2, HiL-Tracking + SAM 2 Video, and FLAIR on 2 unseen holdout videos of nurse sharks. Object detector methods have near-zero segmentation accuracy on the second video.}
  \label{fig:twovids}
\end{figure*}

\begin{figure*}[h!]
  \centering
  \includegraphics[width=\linewidth]{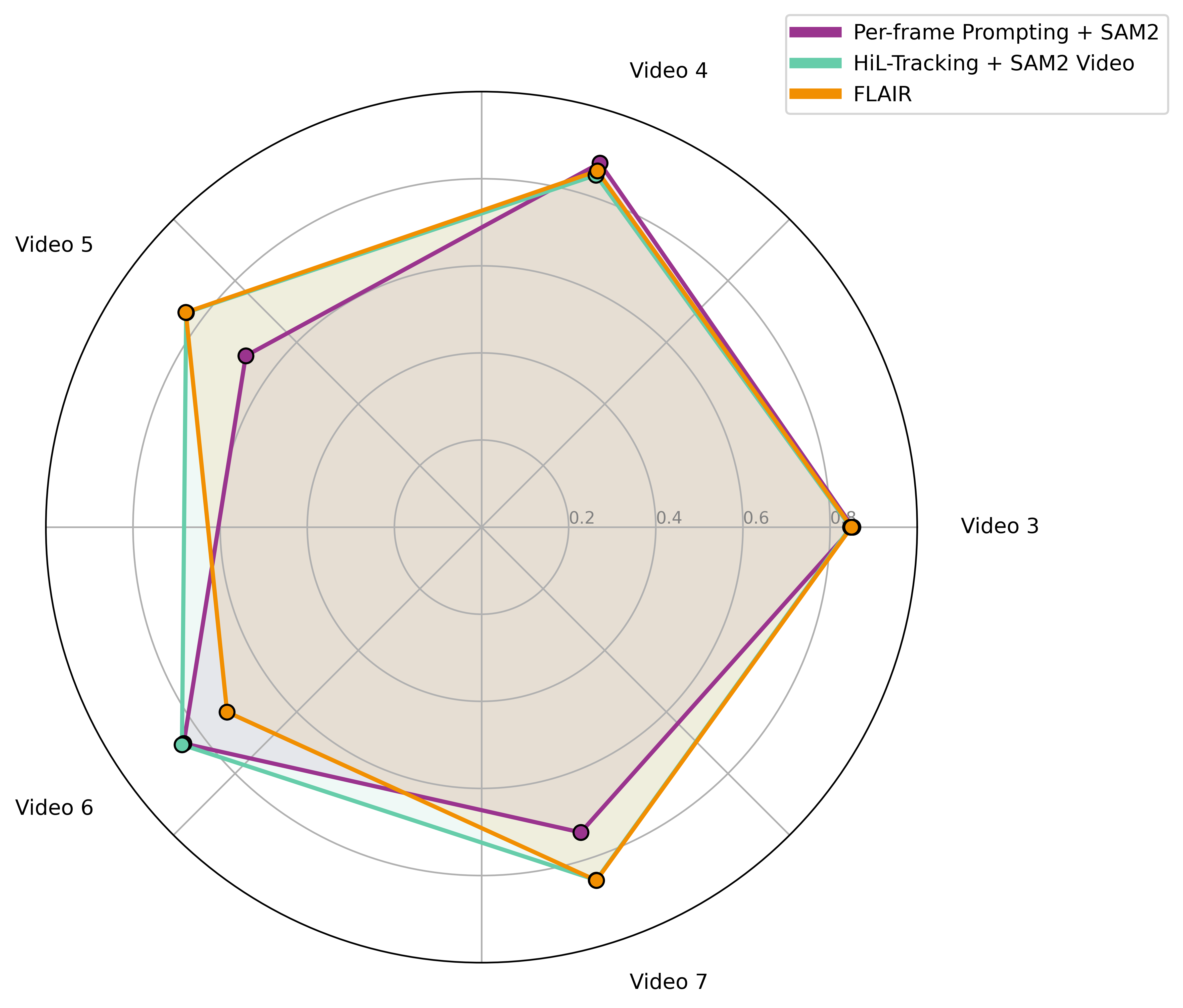}
  \caption{Dice score comparison of Per-frame Prompting, HiL Tracking, and FLAIR on 5 videos containing nurse sharks. FLAIR has competitive performance with both methods that require a human in the loop.}
  \label{fig:radar}
\end{figure*}

\subsection{Shark Segmentation}
To better evaluate performance across models, segmentation metrics were measured on predicted masks. Segmentation is a more relevant task to evaluate than basic object detection for downstream pipelines, including studying biometrics and biological interactions. Dice score was used to evaluate segmentation performance, which measures the spatial overlap between the predicted and ground truth masks, with a value of 1 indicating perfect agreement and 0 indicating no overlap. All models had relatively high Dice scores for the first holdout video, that was deemed to be "in distribution" of the other videos, as it was taken on the same day and location as two of the training videos. However, both object detectors (YOLOv8 and DETR) were unable to identify bounding boxes in the second video, which was taken on a different day and weather conditions, resulting in low Dice scores of 0.014 and 0. The other semi- and fully-autonomous approaches (Per-frame Prompting, HiL-Tracking, FLAIR) had high Dice scores of 0.780, 0.839, and 0.837 respectively, on this second video that was deemed to be out of distribution—a stronger test of the models' generalizability (Table \ref{tab:all_methods_vids}).

The Per-frame Prompting and HiL-Tracking approaches both had accurate segmentation mask predictions on the other 5 videos, with mean Dice scores of 0.796 and 0.847 respectively, (Table \ref{tab:all_methods_vids}). HiL-Tracking outperforming the Per-frame Prompting highlights the advantage of using video understanding in object segmentation, having slightly higher segmentation accuracy while requiring significantly lower human effort. FLAIR achieved comparable Dice scores on all 5 of the videos as seen in Table \ref{tab:all_methods_vids}, with a mean Dice score of 0.825. FLAIR outperformed Per-frame Prompting and HiL-Tracking on Videos 5 and 7. 

\subsection{Biometrics Case Study}

\begin{figure*}[h!]
  \centering
  \includegraphics[width=\linewidth]{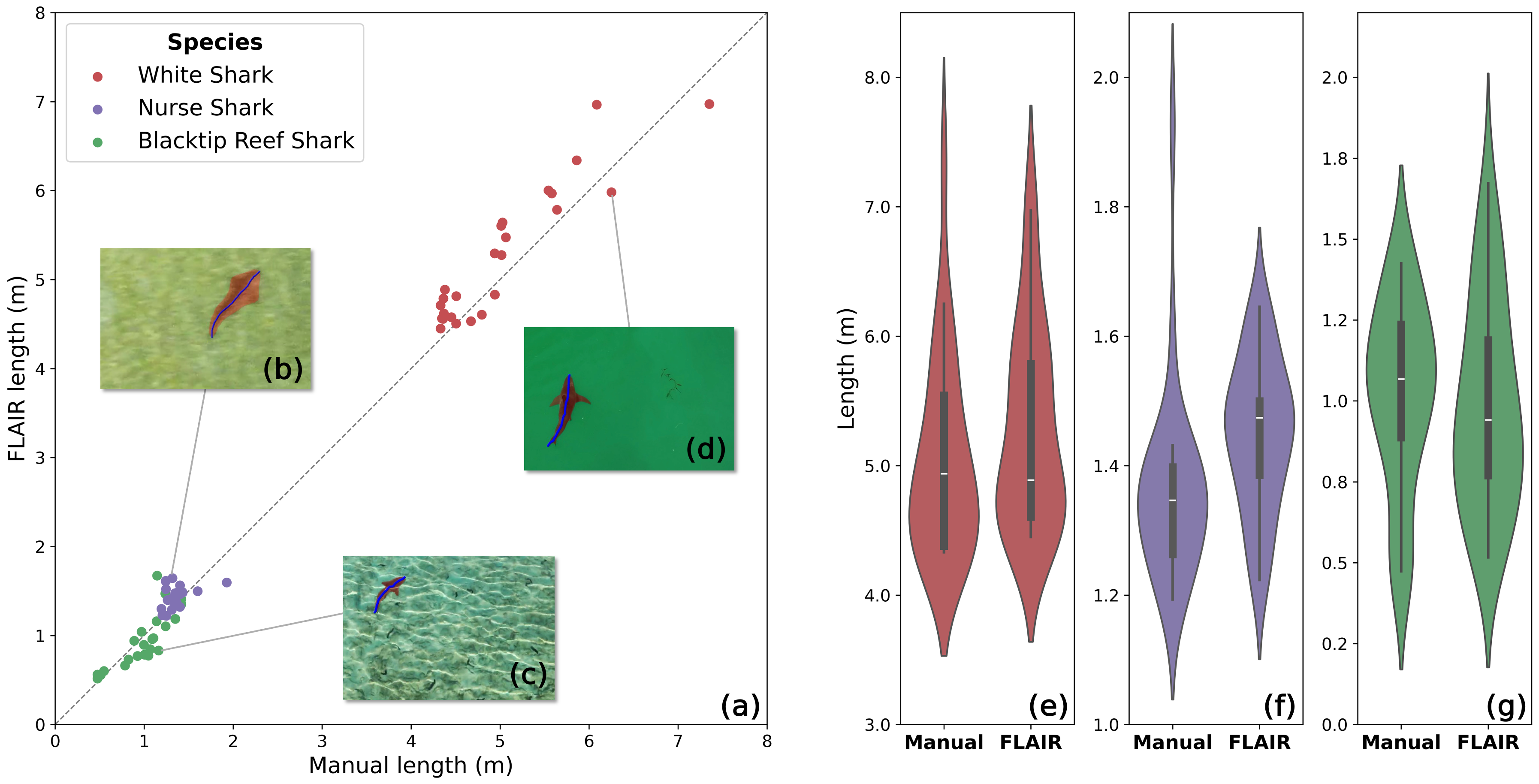}
  \caption{(a) Comparison between manual length measurements and automated length calculation from FLAIR masks across three shark videos for (b) Pacific nurse shark, (c) blacktip reef shark, and (d) white shark. Violin plots show the distribution of length measurements measured manually and calculated from FLAIR masks from sampled frames across white shark (e), Pacific nurse shark (f) and blacktip reef shark (g) videos. Internal box plots show the median (white horizontal line), interquartile range (25th-75th percentile, grey rectangles) and the grey vertical lines extend to the minimum and maximum.}
  \label{fig:length}
\end{figure*}

Body length and tailbeat frequency (TBF) were calculated from FLAIR masks for sampled frames from three individual videos of a white shark, a Pacific nurse shark, and a blacktip reef shark. These calculations were compared to manual measurements of body length and TBF. The body lengths derived from FLAIR-predicted masks (reported as mean ± standard deviation) for the white shark, Pacific nurse shark, and blacktip reef shark, were 5.3 ± 0.8 m, 1.5 ± 0.1 m, and 1.0 ± 0.3 m, respectively. Similarly, the body lengths derived from manual annotation for the white shark, Pacific nurse shark, and blacktip reef shark, were 5.0 ± 0.8 m, 1.4 ± 0.1 m, and 1.0 ± 0.3 m, respectively. The predicted body lengths are nearly identical to the manually measured lengths, as these values are tightly concentrated around the line $y = x$ in Fig. \ref{fig:length}a. The distributions of predicted body length across all three sharks are also highly similar to ground truth measurements (Figs \ref{fig:length}e-\ref{fig:length}g). In addition, visual inspection of masks and predicted centerlines shows accurate segmentation and center line estimation of the sharks (Figs. \ref{fig:length}b-\ref{fig:length}d). 

\begin{figure*}[h!]
  \centering
  \includegraphics[width=\linewidth]{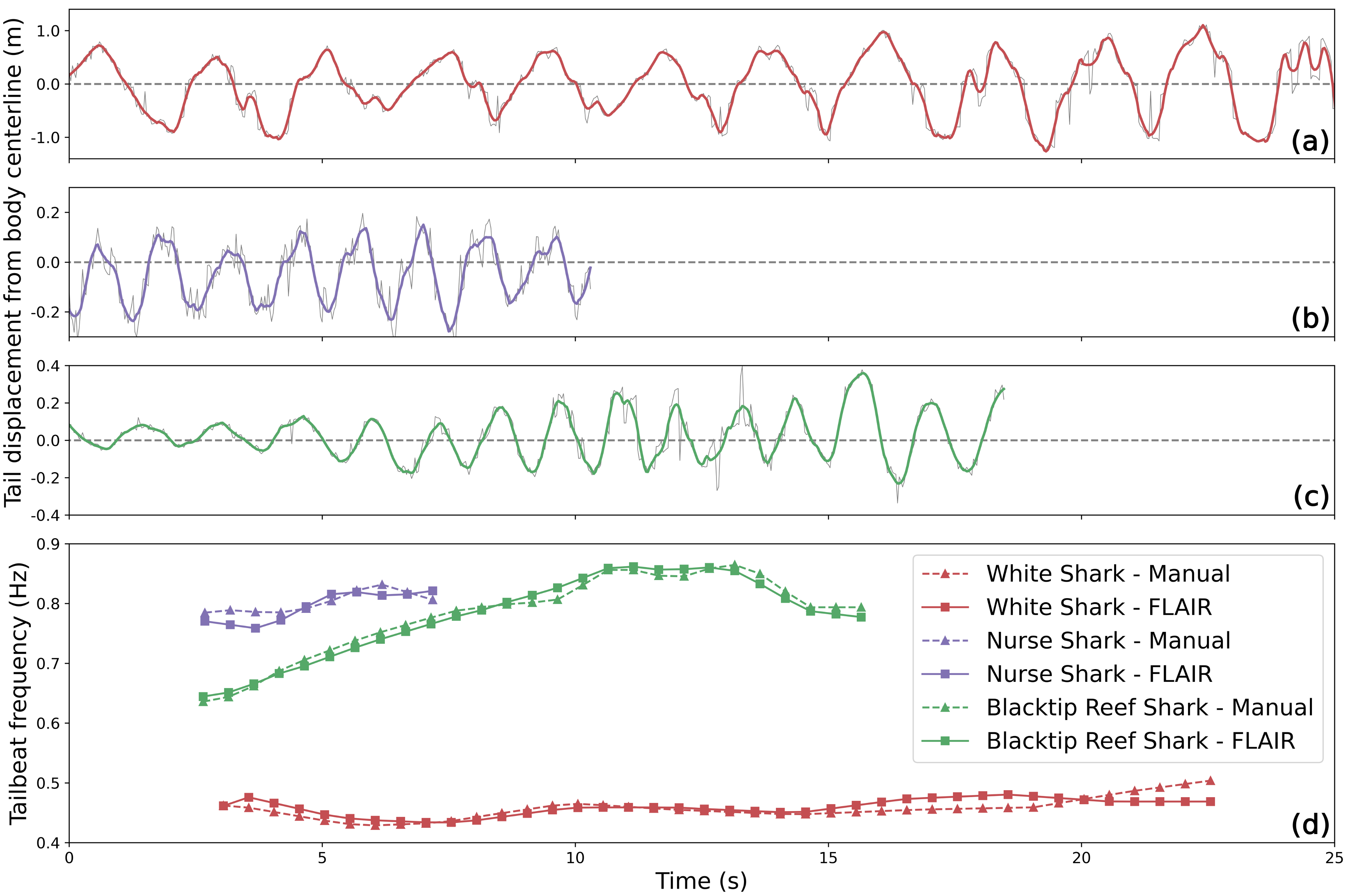}
  \caption{Tail displacement from body centerline over time for white shark (a), Pacific nurse shark (b), and blacktip reef shark (c), derived from FLAIR masks. Thin grey lines denote the raw signal, while thick colored lines denote the smoothed signal. (d) Comparison between automatically calculated tailbeat frequency (from FLAIR masks) and manually measured tailbeat frequency for all 3 species of shark.}
  \label{fig:tbf}
\end{figure*}

Tail beat frequency was computed following processing and smoothing of raw signal of tail displacement from the center line as shown in Figs \ref{fig:tbf}a-\ref{fig:tbf}c. Smoothing the signal reduced noise and improved centerline crossing calculation. TBF predicted from FLAIR masks very closely followed manually calculated TBF, with a mean error of 2.07\% across all three species. All predicted TBF measurements were within 7\% error from the manual measurements. TBF was relatively constant for nurse and white sharks across time, but increased in the blacktip shark across the video, peaking at 0.86 tail beats per second, corresponding to 1.16 seconds per tail beat. (Fig. \ref{fig:tbf}d).  

\subsection{Efficiency Comparisons}
To compare efficiency and speed between methods, we measured the time required for labeling bounding boxes and segmentation masks across the video dataset. We found that manual labeling took an average of 10.5 seconds per frame for bounding box annotation and an additional 45 seconds per frame for mask segmentation for a total of 55.5 seconds per frame. For both the object detection methods and Per-frame Prompting, the human effort time was 10.5 seconds per frame for the bounding box annotation. HiL-Tracking requires 0.15 seconds per frame for human annotation (including watching the video and selecting bounding boxes) and FLAIR requires no human effort. To put this in context, the human effort time for a 5 minute aerial drone video at 30 FPS requires approximately 139 hours for manual labeling, 26 hours for object detection and Per-frame Prompting, 22.5 minutes for HiL-Tracking and 0 minutes for FLAIR. 

\section{Discussion}
In this study, we present FLAIR, a fully automatic semantic segmentation strategy that markedly improves the efficiency of studying marine animals in aerial drone videos. We found FLAIR performed better than several state-of-the-art (SoTA) object detection models used to prompt SAM 2. This demonstrates that traditional object detection methods, such as YOLOv8 and DETR, require larger and more diverse training sets for appropriate generalizability across videos with different conditions, which are often not available for data-poor species. Training these models also requires significant annotator effort. Furthermore, FLAIR performed competitively with several approaches requiring a human in the loop (while FLAIR is fully automated). In addition, FLAIR and HiL-Tracking segmented sharks accurately across a diverse set of videos containing Pacific nurse sharks, blacktip reef sharks, and white sharks. Given the scarcity of diverse spatial and temporal labeled aerial drone data in the study of marine animals, FLAIR presents an efficient and accessible method for a wide range of applications. 

We also show that FLAIR segmentations can be used for biometric analysis of shark imagery, including computation of animal length—which is essential for understanding shark population demographics. Shark speed, tail beat frequency, and other kinematic measurements can also be accurately estimated using FLAIR.  Estimated lengths were accurate across diverse conditions, including varying camera angles and body poses (Figs \ref{fig:length}b-\ref{fig:length}d). These kinematics estimates from aerial imagery can be monitored across time and habitat changes, and studied in combination with biological metadata to assess predator-prey interactions, cooperative behavior \cite{whitney2010identifying}, and much more \citep{gleiss2009multi}. These observations are particularly relevant for understanding marine animal behavior and physiology on a fine scale, and allow biologists to quantitatively study the energetics and movement patterns of large marine animals \citep{andrzejaczek2019biologging}. 

The primary limitation for both the semi- and fully-automated segmentation tracking methods is the accuracy of the segmentations, which can have downstream effects on length and width predictions for animals. When the shark is swimming near the seafloor in very shallow water, SAM 2 will occasionally segment the shadow of the shark along with the shark itself. In turbid water, the pectoral and caudal fins are occasionally left out of the segmentation, or the caudal lobe will be segmented separately from the rest of the body. In addition, obtaining precise biometrics from segmentation masks may be challenging in aerial videos where drone metadata isn't available. Ultimately, the quality of the drone aerial imagery has a significant effect on the accuracy of detection and segmentation for any marine and terrestrial species \citep{ramos2022drone}. 

The core advantage of FLAIR is its potential to generalize to new marine and terrestrial aerial datasets. This pipeline is directly applicable to the tracking of any animal or object from aerial imagery. FLAIR is particularly well-suited for ecology and conservation, where high-quality data are often limited and object tracking plays a crucial role \citep{weinstein2018computer}. Using aerial imagery to study wildlife allows for non-invasive tracking and observation, capturing dynamic information about animal biomechanics, interactions, and behaviors \citep{Torres2022, hansen2022mechanisms, gray2019drones, bierlich2024automated}. We hope that this framework will be applied across diverse ecosystems and species, providing a scalable solution for addressing conservation challenges, informing policy, and fostering sustainable management practices in both marine and terrestrial environments \citep{buchelt2024exploring, stark2018combining}.

Deep learning is transforming ecological research by enabling scientists to process and analyze massive datasets of wildlife imagery. Integrating state-of-the-art foundation model frameworks to derive biological conclusions and understand fine-grained changes in ecosystems should be a priority. The methods presented here, namely FLAIR, allow for completely automated detection and tracking of animals from aerial imagery, along with streamlined pipelines for downstream biometrics. As foundation models become increasingly powerful and efficient, we expect that methods like FLAIR will be scalable tools for understanding complex ecological interactions.

\begin{ack}
    Many thanks to M. Lara, S. Lara and A. Lara from Dive Center Cuajiniquil and C. Lowe for supporting data collection as a joint project between UCR CIMAR and CSULB Shark Lab. Our sincere thanks are extended to J. Meribe for his assistance with dataset labeling. We also gratefully acknowledge funding from the Save Our Seas Foundation (Project No. 664), the American Elasmobranch Society, and the Richard D. Greene Research Fellowship, which supported many of our drone survey trips. Support for data collection was also provided by the Mel Lane Student Grants Program from the Stanford Woods Institute for the Environment and Stanford ASSU.
\end{ack}

\section{Data Availability}
% TODO: update data availability
    All code for data processing, training and testing neural networks, and generating figures is available in this Github repository [redacted for review]. The datasets generated and analyzed during the current study are available in this Zenodo archive [redacted for review] or from the corresponding author on reasonable request. 

\section{Conflict of Interest}\=
    The authors declare no conflict of interest. 

\section{Author Contributions}
    This study is a collaboration between authors from multiple countries, including scientists based in Costa Rica, where the study was conducted. Each author participated from the beginning of the research process, ensuring that the diverse perspectives they represented were considered.
    
    C.K.L., M.E.L., and J.V.C. conceived the ideas and designed methodology; S.M.M and M.E. collected the data; C.K.L. led development of software, C.K.L, M.E.L., and J.V.C. analysed the data; C.K.L. led the writing of the manuscript. All authors contributed critically to the drafts and gave final approval for publication.

\bibliographystyle{plainnat}  
\bibliography{main}

\end{document}